# Efficient enumeration of instantiations in Bayesian networks

**Sampath Srinivas**
Microsoft Corporation
One Microsoft Way
Redmond, WA 98052
sampaths@microsoft.com

**Pandurang Nayak**
Recom Technologies
NASA Ames Research Center, MS 269-2
Moffett Field, CA 94035
nayak@ptolemy.arc.nasa.gov

## Abstract

Over the past several years Bayesian networks have been applied to a wide variety of problems. A central problem in applying Bayesian networks is that of finding one or more of the most probable instantiations of a network. In this paper we develop an efficient algorithm that incrementally enumerates the instantiations of a Bayesian network in decreasing order of probability. Such enumeration algorithms are applicable in a variety of applications ranging from medical expert systems to model-based diagnosis. Fundamentally, our algorithm is simply performing a lazy enumeration of the sorted list of *all* instantiations of the network. This insight leads to a very concise algorithm statement which is both easily understood and implemented. We show that for singly connected networks, our algorithm generates the next instantiation in time polynomial in the size of the network. The algorithm extends to arbitrary Bayesian networks using standard conditioning techniques. We empirically evaluate the enumeration algorithm and demonstrate its practicality.

## 1 INTRODUCTION

Over the last several years Bayesian networks have been applied to a wide variety of problems ranging from medical diagnosis [Heckerman *et al.*, 1992; Horvitz *et al.*, 1988] and natural language understanding [Charniak and Goldman, 1991] to vision [Levitt *et al.*, 1989] and map learning [Dean, 1990]. A central problem in such applications is to use the network to generate explanations for observed data. Such explanations correspond to *instantiations* of the network (i.e., value assignments to each node in the network) with the structure of the network providing the explanation for the values. Each such instantiation has an associated probability that can be computed from the specification of the Bayesian network (see [Pearl, 1988a] for details). Hence, finding one or more of the most probable instantiations of a Bayesian network is a problem of central importance.

Pearl [1987] has developed an elegant message-passing algorithm for computing the most probable instantiation of a Bayesian network. This algorithm runs in polynomial time on singly connected networks, and can be extended to arbitrary networks through conditioning. It can also be used to compute the second most probable instantiation. Dawid [1992] has also developed an algorithm that computes the most likely instantiation using the junction tree of the Bayesian network. This algorithm is inherently applicable to arbitrary networks.

While generating the most probable instantiation is important, it is inadequate for a variety of applications. Instead, such applications require that different instantiations of the network be *enumerated* in decreasing order of probability. For example, in a medical diagnosis application, the most probable diagnosis is rarely adequate; doctor's typically want the *differential diagnosis*, i.e., the set of plausible diagnoses that can explain the observed symptoms. The differential diagnosis is used in two ways: (a) to decide upon a set of tests that best distinguish between the various diagnoses; and (b) to help design a treatment plan, e.g., to select a plan that as applicable to all (or most) of the possible diagnoses. In Bayesian network terms, each diagnosis corresponds to an instantiation and a differential diagnosis is generated by enumerating instantiations in decreasing order of probability.

Turning to another field, Hidden Markov Models (HMMs) have been used to find the most likely labeling of words with parts of speech in natural language applications. The standard HMM model used in this application [Charniak *et al.*, 1993] can be repre-sented as a singly connected Bayesian network. Each labeling of words with parts of speech corresponds to an instantiation of the network. Current techniques compute the most likely labeling, i. e., the most likely instance of the corresponding Bayesian network. However, it may be happen that the most likely labeling is rejected by the semantic analysis phase of the natural language system. In such a situation, the next most likely labeling is necessary. Enumerating labelings in decreasing order of probability corresponds directly to the problem of enumerating instantiations of the Bayesian network in decreasing order of probability.

Finally, enumerating instantiations of Bayesian networks is also needed to extend model-based diagnosis to handle dependent component failures. Some of the best model-based diagnosis algorithms [de Kleer and Williams, 1989; de Kleer, 1991] are based on enumerating candidates in decreasing order of prior probability, and checking these candidates for consistency with the observations.[1] The enumeration algorithms used to date make the strong assumption that component failures are mutually independent. Dependent component failures can be represented using a Bayesian network in which the nodes represent components and node values represent component modes, so that network instantiations correspond to candidates. Existing model-based diagnosis algorithms can therefore be extended to handle dependent component failures using an algorithm to generate network instantiations in decreasing order of probability (see [Nayak and Srinivas, 1995]).

In this paper we develop an efficient algorithm to enumerate instantiations of a Bayesian network in decreasing order of probability. Our algorithm can be viewed as a generalization of Pearl's message passing algorithm for generating the most probable instantiation. We develop our algorithm in two phases. In the first phase, described in Section 2, we develop an algorithm to generate the *entire* list of network instantiations, sorted in decreasing order of probability. Of course, generating the entire list of network instantiations is impractical since the number of instantiations is exponential in the size of the network. Hence, in the second phase, described in Section 3, we show how the above algorithm is modified to *incrementally* compute one instantiation at a time in decreasing order of probability. We analyze the complexity of the incremental algorithm in Section 4, and show that for singly connected networks the next instantiation can always be generated in time polynomial in the size of the network. In Section 5 we consider extensions to multiply connected networks and to evidence nodes. Section 6 discusses experimental results from our implementation of the algorithms and demonstrates its practicality. Section 7 discussed related work. We conclude in Section 8 with a discussion of future work.

---

[1] A candidate is an assignment of nominal or failure modes to components.

## 2  COMPUTING THE ENTIRE LIST OF INSTANTIATIONS

Pearl [1987] describes a message passing algorithm for computing the most likely instance of a singly connected Bayesian network. Our enumeration algorithm is also a message passing algorithm, and can be viewed as a generalization of Pearl's algorithm. It operates as follows. An arbitrary node in the Bayesian network is chosen as the starting node. The starting node requests all its neighbors for messages pertaining to the computation of a list of instances sorted in decreasing order of probability. These messages pertain to instantiations of the part of the network reachable through the neighbor. When the starting node has received the messages it combines them appropriately and returns the entire list of instantiations of the Bayesian network sorted in decreasing order of probability. When a neighbor is requested to give a message, it recursively requests each of its neighbors (except for the original requesting node) for a message. It combines these messages appropriately and passes them on to the requesting node. As we will see, the independence properties of the singly connected network make such a message passing algorithm possible.

The description of the message passing algorithm thus reduces to the description of the operations at a single node. The description explains what the messages are and how the messages coming from neighbors are combined and sent to the requesting node. As noted earlier, we start by describing how to compute the *entire* list of instantiations in decreasing order of probability. In the next section we show how to modify this algorithm to make it compute one instance at a time (on demand).

### 2.1  WHAT ARE THE MESSAGES?

We now define the messages sent between nodes. We start by defining some terminology. Consider two nodes $A$ and $B$ that are connected by an arc in a Bayesian network $R$. We use $R_{A\|B}$ to refer to the set of all the nodes in the subnetwork containing $A$ when the arc connecting $A$ and $B$ is disconnected. Note that the arc between $A$ and $B$ can be in either direction.

Suppose that node $X$ requests node $Y$ for a message, and let $Y$ be a *parent* of $X$. We will refer to the mes-sage that $Y$ sends $X$ as $\pi^l_{Y \to X}$.[2] The direction of the arrow in the subscript refers to the direction of the message (from $Y$ to $X$) and not the direction of the arc in the Bayesian network. The superscript $l$ (for "list") reminds us that the message is being used to compute the ordered list of all instances of the Bayesian network.

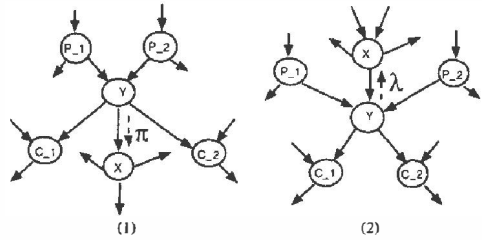

Figure 1: Message passing – two possible cases.

$\pi^l_{Y \to X}$ is a vector indexed by the states $y$ of $Y$. The location $\pi^l_{Y \to X}[Y = y]$ contains a list of all instantiations $i$ of the nodes in $R_{Y \| X}$ such that $Y$ has state $y$ in $i$. The list elements are arranged in decreasing order of probability. The probability is stored with each list element. Since $R$ is a singly connected network, the elements in $R_{Y \| X}$ form a complete Bayesian network in themselves. Hence it is possible to compute the probability of each instantiation of $R_{Y \| X}$ without regard to $X$ or any node reachable through $X$.

Now consider the case where $Y$ is a *child* of the requesting node $X$ in the Bayesian network. We will refer to the message that $Y$ sends $X$ as $\lambda^l_{Y \to X}$. The $\lambda^l_{Y \to X}$ message is indexed by the states $x$ of $X$. $\lambda^l_{Y \to X}[X = x]$ contains a list of instantiations $r$ of $R_{Y \| X}$ sorted by decreasing order of the probability $P(R_{Y \| X} = r | X = x)$. This is the conditional probability of the instance $r$ given $X$ is in state $x$. The probability is stored with each list element.

Note that observing the value of $X$ makes the nodes in $R_{Y \| X}$ independent of all nodes in $R_{X \| Y}$. Hence given a state $x$ of $X$ and an instantiation $r$ of $R_{Y \| X}$, it is possible to compute the probability $P(R_{Y \| X} = r | X = x)$ locally within the subnetwork formed by the nodes in $R_{Y \| X}$.

## 2.2  COMPUTING THE MESSAGES

Suppose that node $X$ has requested node $Y$ for a message. We describe the computations that $Y$ performs in computing the message.

$Y$ first recursively asks for messages from all its neighbors (except for $X$). After they are available it computes the message meant for $X$. There are two cases: $Y$ is either a parent or a child of $X$.

### 2.2.1  $Y$ is a parent of $X$

Consider an example of the first case (Figure 1.1). Say we are given an instance $r_{P_1 \| Y}$ of the set $R_{P_1 \| Y}$, with $P_1 = p_1$ in $r_{P_1 \| Y}$. Similarly, say we are given an instance $r_{P_2 \| Y}$ of $R_{P_2 \| Y}$, with $P_2 = p_2$ in $r_{P_2 \| Y}$. Furthermore, let $r_{C_1 \| Y}$ and $r_{C_2 \| Y}$ be any two instances of $R_{C_1 \| Y}$ and $R_{C_2 \| Y}$.

If we append all these instances together and add in a choice of of state for $Y$, say $Y = y$, we get a full instance $r_{Y \| X}$ of $R_{Y \| X}$. The independence properties of a singly connected Bayesian network implies that:

$$\begin{aligned} P(r_{Y \| X}) &= P(Y = y | P_1 = p_1, P_2 = p_2) \times \quad (1) \\ &\quad P(r_{P_1 \| Y}) P(r_{P_2 \| Y}) \times \\ &\quad P(r_{C_1 \| Y} | Y = y) P(r_{C_2 \| Y} | Y = y) \end{aligned}$$

Note that $r_{Y \| X}$ is an element of $\pi^l_{Y \to X}[Y = y]$. Similarly $r_{P_1 \| Y}$ is an element of $\pi^l_{P_1 \to Y}[P_1 = p_1]$ and $r_{P_2 \| Y}$ is an element of $\pi^l_{P_2 \to Y}[P_2 = p_2]$. In addition, $r_{C_1 \| Y}$ is an element of $\lambda^l_{C_1 \to Y}[Y = y]$ and $r_{C_2 \| Y}$ is an element of $\lambda^l_{C_2 \to Y}[Y = y]$. The probabilities required in Equation 1 are exactly those stored with these elements (see Section 2.1).

Figure 2 shows the algorithm that uses Equation 1 to compute the message $\pi^l_{Y \to X}$. The following terminology is used in this algorithm. Given two ordered lists of instances $L_1$ and $L_2$ ordered in decreasing order of probability, let $L_1 \otimes L_2$ be the ordered list composed of all possible combinations of the instances where one element is chosen from $L_1$ and one element is chosen from $L_2$. The probability of the combination is the product of the stored probabilities of the components. Given an ordered list of instances $L$ and a number $k$, let $k \odot L$ be the list where the probability of every instance in $L$ is multiplied by $k$. Finally, given a list of ordered instance lists $LL$, let $Merge(LL)$ be the list formed by merging the constituent lists of $LL$ into a single ordered list. Each of the constituent lists of $LL$ is assumed to contain instances of the same set of variables.

The discussion above leads directly to the algorithm in Figure 2. This algorithm computes $\pi^l_{Y \to X}$ from the messages coming to it from $P_1$, $P_2$, $C_1$ and $C_2$. In essence, for each state $y$ of $Y$, the algorithm is generating every element of $\pi^l_{Y \to X}[Y = y]$ and ensuring that the elements are put together into a list in decreasing order of probability. The algorithm is easily

---

[2] We follow Pearl in choosing $\pi$ and $\lambda$ as the message names.

---

**begin Compute-$\pi^l_{Y \to X}(Y, X)$**
For all states $y$ of $Y$:

1. $L^\lambda_y = \lambda^l_{C_1 \to Y}[Y = y] \otimes \lambda^l_{C_2 \to Y}[Y = y]$

2. Initialize $LL$ to the empty list. $LL$ is a list of lists.

3. For all combinations $< p_1, p_2 >$ of the states of $P_1$ and $P_2$:

    (a) Let $k = P(Y = y | P_1 = p_1, P_2 = p_2)$.
    (b) $L = k \odot \pi^l_{P_1 \to Y}[P_1 = p_1] \otimes \pi^l_{P_2 \to Y}[P_2 = p_2] \otimes L^\lambda_y$
    (c) Add $L$ to $LL$.

4. $\pi^l_{Y \to X}[Y = y] = Merge(LL)$

**end Compute-$\pi^l_{Y \to X}$**

Figure 2: Algorithm when $Y$ is a parent of $X$.

---

**begin Compute-$\lambda^l_{Y \to X}(Y, X)$**
For all states $x$ of $X$:

1. Initialize $LL$ to the empty list. $LL$ is a list of lists.

2. For all states $y$ of $Y$:

    (a) $L^\lambda_y = \lambda^l_{C_1 \to Y}[Y = y] \otimes \lambda^l_{C_2 \to Y}[Y = y]$
    (b) For all combinations $< p_1, p_2 >$ of the states of $P_1$ and $P_2$:
        i. Let $k = P(Y = y | P_1 = p_1, P_2 = p_2, X = x)$.
        ii. $L = k \bullet \pi^l_{P_1 \to Y}[P_1 = p_1] \otimes \pi^l_{P_2 \to Y}[P_2 = p_2] \otimes L^\lambda_y$
        iii. Add $L$ to $LL$.

3. $\lambda^l_{Y \to X}[X = x] = Merge(LL)$

**end Compute-$\lambda^l_{Y \to X}$**

Figure 3: Algorithm when $Y$ is a child of $X$.

---

adapted to the case where $Y$ has an arbitrary number of parents and an arbitrary number of children.

#### 2.2.2  $Y$ is a child of $X$

Now consider the case where $Y$ is a child of $X$. This situation is shown in Figure 1.2. The same argument used in the first case leads to the algorithm shown in Figure 3.[3]

---

[3]At the expense of clarity, this algorithm can be improved by computing and saving $L^\lambda_y$ for all $y$ before entering the main loop.

---

**begin Compute-ordered-instances($B$)**

1. Choose an arbitrary node $R$ of the Bayesian network $B$ and add a dummy node $D$ as parent.

2. **Compute-message**$(R, D)$

3. Return $\lambda^l_{R \to D}[D = \hat{d}]$.

**end Compute-ordered-instances**

---

**begin Compute-message$(Y, X)$**

1. For all neighbors $N$ of requested node $Y$ except the requesting node $X$ do:

    **Compute-message**$(N, Y)$

2. If $Y$ is a parent of $X$ then:

    **Compute-$\pi^l_{Y \to X}(Y, X)$**

    else

    **Compute-$\lambda^l_{Y \to X}(Y, X)$**

**end Compute-message**

Figure 4: Algorithm for computing ordered instance list.

### 2.3  COMBINING THE MESSAGES

The algorithm **Compute-ordered-instances** for computing the ordered list of instances of the entire Bayesian network follows directly from **Compute-$\pi^l_{Y \to X}$** and **Compute-$\lambda^l_{Y \to X}$**.

We choose an arbitrary node $R$ of the Bayesian network as the root node and add a dummy node $D$ as a parent of $R$. $D$ has only one state $\hat{d}$. Hence automatically, $P(D = \hat{d}) = 1$. Let the parents of $R$ before adding $D$ be the set $\mathbf{S}_R$. Let $\mathbf{s}_R$ be a joint state of $\mathbf{S}_R$. Assume that the conditional probability was defined by the table $P_{old}(R|\mathbf{S}_R)$. The conditional probability distribution of $R$ after $D$'s addition is set to be $P_{new}(R = r|\mathbf{S}_R = \mathbf{s}_R, D = \hat{d}) = P_{old}(R = r|\mathbf{S}_R = \mathbf{s}_r)$. This ensures that effectively, $R$ is independent of $D$, and hence if $D$ requests $R$ for a message, then $\lambda^l_{R \to D}[D = \hat{d}]$ contains exactly the list of ordered instances of the entire network. The full algorithm is described in Figure 4.

## 3  COMPUTING ONE INSTANCE AT A TIME

The previous section developed an algorithm that computes the *entire* list of ordered instances. Hence, though it takes full advantage of the independence properties of the network to decompose the problem, it's run time is inherently exponential since the number of instances is exponential. In this section, we modify that algorithm to return one instance at a time from the ordered list. The next instance is computed only on demand, i.e., we make the computation *lazy*.

Specifically, all that is required is to make the computation of the list operations $\otimes$, $\odot$ and *Merge* lazy. The modified **Compute-ordered-instances** returns a *lazy list*. Initially, a lazy list contains only the first element of the list. The rest of the elements are stored as a delayed computation in the list's data structure. Each time we demand the next element, the delayed computation is called. It performs only the necessary computations to compute the next element. This element is added to the end of the list. The computation then delays itself again[4].

Note that the list $LL$ in **Compute**-$\pi^l_{Y \to X}(Y, X)$ and **Compute**-$\lambda^l_{Y \to X}(Y, X)$ is not a lazy list. However, each element in $LL$ is a lazy list. This observation implies that the delayed computations will perform only the list operations $\otimes$, $\odot$ and *Merge*. Let the lazy versions of $\odot$, $\otimes$, and *Merge* be $\odot_z$, $\otimes_z$, and $Merge_z$, respectively.

The definitions of $\odot_z$ and $Merge_z$ is straightforward. Given a constant factor $k$ and a lazy list $L_z$ as argument $\odot_z$ multiplies $k$ into the probability of the first element of $L_z$ and returns it. It then wakes up the delayed computation in $L_z$. This results in the second element of $L_z$ being generated. It then goes to sleep. On the next call it multiplies the constant factor into the second element and returns it. It then generates the third element of $L_z$ and goes to sleep and so on. On each call, it performs $O(1)$ computations (not counting the computation within $L_z$'s delayed computation).

On each call, $Merge_z$ goes through its argument $LL$ looking at the probability of the current element of every lazy list in $LL$. It returns the element $C_{max}$ with maximum probability. Let $L_{max}$ be the list from which $C_{max}$ came. $C_{max}$ is popped off $L_{max}$. $Merge_z$ now wakes up the computation of $L_{max}$ till the next element of $L_{max}$ is generated and this is made the current element of $L_{max}$. It then goes to sleep. On each call, $Merge_z$ performs $O(Length(LL))$ comparisons (not counting the computation within the delayed computation of $L_{max}$).

### 3.1 AN EFFICIENT WAY OF MAKING $\otimes$ LAZY

The operation $\otimes$ takes two *ordered* lists $L_I$ and $L_J$ as arguments and returns an ordered list where each element is a compound element composed of one el-ement from $L_I$ and one element of $L_J$. The numerical value associated with the compound element is the product of the numerical values associated with the constituents. The list which is returned is ordered by this numerical value.

---
[4]See [Charniak *et al.*, 1987] for details on implementing lazy list operations.

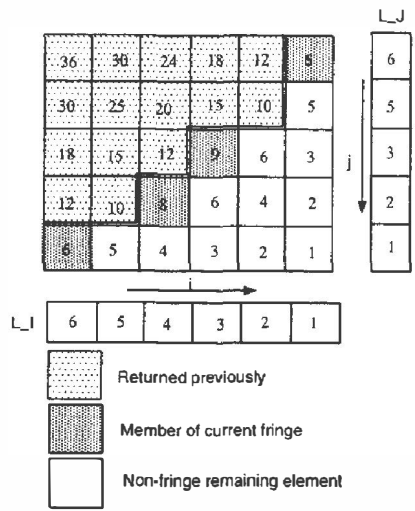

Figure 5: The fringe in $\otimes_z$.

Consider the example shown in Figure 5. List $L_I$ is shown along the rows of a matrix and list $L_J$ is shown along the columns. In this example, the elements of the list are the numerical values themselves. Each location in the matrix is the product of the appropriate elements of $L_I$ and $L_J$.

An element $a(i_2, j_2)$ in the matrix is *dominated* by an element $a(i_1, j_1)$ if it is necessarily less than or equal to $a(i_1, j_1)$ regardless of the actual values in the ordered lists $L_I$ and $L_J$. We see directly that $a(i_2, j_2)$ is dominated by $a(i_1, j_1)$ iff $i_1 \leq i_2$ and $j_1 \leq j_2$. We will call the element $a(i+1, j)$ as the *dominated neighbor along dimension i* of $a(i, j)$. That is, an element's dominated neighbor along a dimension is the element immediately "below" it along that dimension.

We will now describe $\otimes_z$. Every time $\otimes_z$ is called, it returns the next largest element in the matrix. The remaining elements are those elements of the matrix that have not yet been returned during previous calls to $\otimes_z$. $\otimes_z$ encodes the set of remaining matrix elements using the *fringe*, $F$. The fringe consists of those remaining elements that are *not dominated* by any of the other remaining elements (see Figure 5).

Each time $\otimes_z$ is called, it returns the maximum element, $C_{max}$, of the fringe $F$ and then updates the fringe. The fringe update is easily accomplished by the procedure **Update-fringe** shown in Figure 6. Note that this procedure does not explicitly generate the

---

begin **Update-fringe**

1. Choose the max element $C_{max}$ of the fringe $F$ and delete it from $F$.

2. Along each dimension $K$ do:
   Let $C_K$ be the dominated neighbor of $C_{max}$ along $K$. If $C_K$ is not dominated by any element in $F$ then:

   (a) Compute $C_K$ (i.e., actually multiply the probabilities and create the compound element). This computation might require the computation of the next yet uncomputed element of $L_K$. If so, awaken the computation of $L_K$ so that this element is available.

   (b) Add $C_K$ to the set $F$.

3. Return $C_{max}$.

end **Update-fringe**

Figure 6: Updating the fringe in $\otimes_z$.

---

matrix. It simply retains the matrix indices with each element of $F$ to perform domination tests.

We have assumed in the above discussion that $\otimes_z$ takes only two arguments. However, the identical discussion applies if there are $n$ lists given as arguments. Instead of a two dimensional matrix, we have a $n$ dimensional matrix. The **Update-fringe** procedure applies even when there are $n$ arguments. A general implementation that can handle any number of argument lists can be used to compute $(L_1 \otimes_z L_2 \otimes_z \ldots \otimes_z L_n)$ as $\otimes_z (L_1, L_2, \ldots, L_n)$. Such an implementation is immediately applicable in **Compute-**$\pi^l_{Y \to X}(Y, X)$ and **Compute-**$\lambda^l_{Y \to X}(Y, X)$.

Let $L_{result}$ be the entire ordered list which would result if all the elements returned successively by $\otimes_z (L_1, L_2, \ldots, L_n)$ were computed (by repeated calls to the delayed computation). Consider the situation where the first $k$ elements of $L_{result}$ have been computed and the rest are yet uncomputed. We see that every time we update the fringe, we add at most $n$ elements to it. At the start of the computation the fringe consists of exactly 1 element (viz, the first element of $L_{result}$). Hence after $k$ elements of $L_{result}$ have been computed, the size of the fringe is at most $nk$. Examining **Update-fringe**, we see that when computing the $k + 1$st element, we need $O(nk)$ comparisons to determine $C_{max}$. In addition, we need to make $O(nk)$ domination tests along each of the $n$ dimensions in Step 2. Each domination tests requires $n$ index comparisons. Hence the $k + 1$st element element of $L_{result}$ can be computed with $O(n^3 k)$ comparisons (not counting any operations resulting from waking up computations in any argument list). We note that this is a loose bound.

In practice, as we shall see later, $\otimes_z$ does much better.

## 4  COMPLEXITY OF THE FULL LAZY ALGORITHM

We now compute an upper bound on the complexity of the full lazy algorithm, i.e., the complexity of generating the $k$th most probable instance of the Bayesian network.

Consider a node $Y$ which is computing a message to be sent to node $X$ using **Compute-message**$(Y, X)$. Let $Size(Y)$ be the size of the conditional probability table of $Y$. That is, $Y$ is the product of the cardinalities of $Y$ and each of its parents. Let $Degree(Y)$ be the number of neighbors of $Y$, i.e., the sum of the number of parents and number of children.

We examine the complexity of computing the message where each message element is a lazy list. Specifically, we look at the total complexity of computing the next element in each of these lazy lists. We consider only the computations performed within $Y$, i.e., we exclude the comparisons performed in recursive calls to **Compute-message**.

Examining **Compute-**$\pi^l_{Y \to X}(Y, X)$ and **Compute-**$\lambda^l_{Y \to X}(Y, X)$, we note that the $Merge_z$ and $\odot_z$ operations together perform $O(Size(Y))$ operations.[5]

Now consider the number of operations performed by the $\otimes_z$. Say we have generated the first $k$ elements of every message element list and are looking to generate the $k + 1$st element of each of these lists. We see that the number of operations performed by $\otimes_z$ is bounded by $O(Size(Y) Degree(Y)^3 k)$.

Given a Bayesian network $B$ let $Size(B) = \Sigma_{Y \in B} Size(Y)$. We see that $Size(B)$ measures the amount of information required to specify the network. Let $MaxDegree(B) = \max_{Y \in B} Degree(Y)$.

Say we have generated the $k$ most probable instances of the Bayesian network and are now computing the $k + 1$th most probable instance. We see that $\odot_z$ and $Merge_z$ together perform $O(Size(B))$ operations. $\otimes_z$ performs $O(\,Size(B)\,MaxDegree(B)^3\,k\,)$ operations.

Thus, the overall complexity of generating the $k + 1$st most probable instance is $O(\,Size(B)\,MaxDegree(B)^3\,k\,)$. Note that this is a loose upper bound. There are two reasons. The first is that the bound that we computed earlier on $\otimes_z$ is loose. The second is that we are assuming that every delayed list will be forced to compute its next element in the process of computing the next most probable instance of the entire network. This need not be true. In practice, the algorithm runs much faster (as described later).

---

[5] In this analysis, we consider both comparisons and multiplications as elementary operations.

Table 1: Run times for **Compute-ordered-instances**.

| Time to generate next instance | |
|---|---|
| Number of Bayes net variables | 300 |
| Max. numb. of states per node | 5 |
| $MaxDegree$ | 5 |
| Number of instances generated | 600 |
| Setup time | 11 secs |
| Max. time | 34 msec |
| Min. time | 0 msec |
| Avg. time | 7.8 msec |

| Time to generate next instance | |
|---|---|
| Number of Bayes net variables | 500 |
| Max. numb. of states per node | 6 |
| $MaxDegree$ | 6 |
| Number of instances generated | 600 |
| Setup time | 2 mins |
| Max. time | 50 msec |
| Min. time | 0 msec |
| Avg. time | 11.00 msec |

We also note that when $k = 1$ the algorithm computes the most probable instance of the network. This is exactly what is computed by [Pearl, 1987].

## 5  MULTIPLY-CONNECTED NETWORKS AND EVIDENCE NODES

The algorithm we have presented so far can handle only singly connected Bayesian networks. When Bayesian networks are not singly connected, there is a general scheme called conditioning which can be used to adapt the singly connected algorithms to perform Bayesian network computations [Pearl, 1988b].

Conditioning chooses a set of nodes in the Bayesian network such that observing the values of nodes leaves the resulting network singly connected. This is in accordance with the independence semantics of Bayesian networks. The set of conditioning variables is called the *cutset*. A computation is performed for every possible joint instance of the cutset using the singly connected algorithm and these computations are then combined. In general, domains suitable for modeling with Bayesian networks have a large number of independences and so the size of the cutset is small.

Our algorithm can be adapted directly to handle multiply connected networks using conditioning. For every joint instance $c$ of the cutset, we compute an ordered list of instances $L_c$ of the network. Each element $e_c$ of $L_c$ will be a full network instance. Each element $e_c$ will necessarily have each of the conditioning variables in the state specified by $c$. The probability stored with $e_c$ will be $P(e_c|c)$. $L_c$ can be computed with the algorithm we have developed above. For each list $L_c$ we then compute $L'_c = P(c) \odot L_c$. Here $P(c)$ is the prior probability of the cutset instance $c$. The lists $L'_c$ (one for each cutset instance $c$) are then merged to give the list of all instances in decreasing order of probability.

Let the cutset of Bayesian network $B$ be $\mathbf{C}_B$. Let $Size(\mathbf{C}_B)$ be the size of the joint state space of the variables in the cutset. A loose upper bound for generating the $k + 1$st most probable instance of the Bayesian network is then $O(Size(\mathbf{C}_B)Size(B)MaxDegree(B)^3 k)$. Thus, the $k+1$st most probable instance can be generated in linear time. We note however, that the problem of computing the most likely instance of a Bayesian network, in general, is NP hard. In other words, the constant factor of our linear time algorithm can be extremely large (since it depends on the network characteristics). Thus, our enumeration algorithm is practical only for sparsely connected networks, i.e., networks for which $O(Size(\mathbf{C}_B))$ is small.

Finally, note that for simplicity of exposition, our algorithm description has not made any reference to evidence nodes. A very simple change makes the algorithm generate only those instances which are consistent with the evidence. These instances are generated in decreasing order of conditional probability given the evidence. The change is as follows: For each evidence node in the belief network, delete all states except the observed evidence state. Note that the probabilities associated with the generated instances will be prior probabilities (i.e., without conditioning on the evidence). However, the posterior probability and the prior probability of each of the the instances is related by the same constant, viz, the prior probability of the evidence. Hence the instances are generated in the correct order (i.e., in decreasing order of posterior probability given the evidence).

## 6  IMPLEMENTATION RESULTS

The algorithm described in this paper has been implemented in Lisp. The results reported below are for unoptimized compiled code in Allegro Common Lisp on a Sun Sparcstation 10. The run times reported are milliseconds of CPU time usage.

Run times for **Compute-ordered-messages** are shown in Table 1. The algorithm is implemented on top of IDEAL, a software package for Bayesian network inference [Srinivas and Breese, 1990]. The times shown are for two randomly generated singly connected belief networks. Given the number of nodes $n$, we generated a singly connected Bayesian network nodes with $n$ nodes. The maximum number of neighbors for any node in the network (i.e., the $MaxDegree$ of the network) and the maximum number of states for each node are also specified before the random Bayesian network is generated. The distribution for the belief network is set randomly.

We see that we can compute each instance in the order of tens of milliseconds on the average when the number of nodes is in the order of hundreds. The time to compute instances varies fairly uniformly as the instances are generated. In other words, there is no trend towards increase or decrease in the average time as the number of instances generated increases. We note here that if the algorithm performed in accordance with its worst case analysis there should be a linear increase in run time. In practice, we see that the algorithm does much better.

We note that the time to initialize the algorithm data structures is substantial relative to the time generate instances. Note that the initialization is a one time cost and can be incurred during off-line precomputation.

## 7 RELATED WORK

In addition to its use in explanation, the computation of most likely instantiations of Bayesian networks has been utilized in Bayesian network inference. [Santos and Shimony, 1994] approach the problem of computing marginal probabilities in Bayesian networks by computing the most likely instances which subsume a particular state of a variable and summing over the probability masses of these instances. They formulate the problem of computing the most likely instance as a best first search and also as an integer programming problem. [Poole, 1993] searches through network instantiations to compute prior and posterior probabilities in Bayesian networks. A heuristic search function is used. In the model-based diagnosis community, [de Kleer, 1991] studies a closely related problem – viz, how to focus the diagnostic search on most likely candidates. The common thread in the work discussed above is a best first search through the space of network instantiations – in this paper, we have used the properties of Bayesian networks to reduce the search problem to a direct polynomial algorithm that performs no search.

The work described in this paper is most closely related to the the results presented in [Sy, 1992] and [Li and D'Ambrosio, 1993]. In [Sy, 1992], the author sets up a search for finding the most probable explanation with a particular pruning strategy. The pruning strategy is analyzed and found to yield a polynomial complexity bound for generating the next most probable instance. [Li and D'Ambrosio, 1993] develop an algorithm to compute the next most likely instance by incrementally modifying "evaluation trees" of probability terms. Their algorithm too has a polynomial bound. Our algorithm's complexity is similar to that of [Sy, 1992] and [Li and D'Ambrosio, 1993]. However, in addition, it gives the additional insight that the underlying operation is simply a lazy enumeration of the sorted list of *all* instantiations. This insight leads to a very concise algorithm statement which is both easily understood and implemented.

## 8 CONCLUSION

We have developed an efficient algorithm to enumerate the instantiations of a Bayesian network in decreasing order of probability. For singly connected networks the algorithm runs in time polynomial in the size of the network. An implementation of the algorithm revealed excellent performance in practice. The algorithm has significant applications including explanation in Bayesian network-based expert systems, part-of-speech tagging in natural language systems, and candidate generation (i.e., computing plausible hypotheses) in model-based diagnosis.

As described earlier, our algorithm can be used in model-based diagnosis to generate candidate diagnoses in decreasing order of probability (even when component failures are dependent). The generated candidates are then checked for consistency with observations of the system. We plan to explore a tighter integration of candidate generation and consistency checking. The basic intuition is as follows: When a candidate is found to be inconsistent this gives us information that may allow us infer that some other candidates (which have not yet been generated) are necessarily inconsistent. If this information is fed back to the candidate generator in some way, it can skip enumeration of such candidates. Such pruning has the potential to dramatically improve the overall efficiency of the diagnosis system.

One special case of interest is the situation where the component failures are independent – i.e., a trivial Bayesian network with no arcs. The problem thus reduces to the following: Given a set of discrete variables $X_1, X_2, \ldots, X_n$ and distributions $P(X_i)$, successively compute joint instances in decreasing order of probability. We have developed a linear time algorithm for this special case — each successive instance is computed in $O(n)$. For this special case, we have also developed a tight integration between the candidate generation and consistency checking (along the lines described above). The result is a highly efficient and focused search strategy [Nayak and Srinivas, 1995].

We also plan to explore another very significant application of our algorithm – viz, enumeration of most likely solutions in Constraint Satisfaction Problems.

**Acknowledgements**

We would like to thank the anonymous referees for their feedback and for pointers to some related work.